\documentclass[conference]{IEEE_ITSC}
\IEEEoverridecommandlockouts
\usepackage{graphicx}
\usepackage{subcaption}
\usepackage{amsthm}
\usepackage{amsmath}
\usepackage{color}
\usepackage{amsfonts}
\usepackage{algorithm} 
\usepackage{algpseudocode}
\usepackage{hyperref}
\usepackage{titlesec}
\usepackage{indentfirst}
\usepackage{hyperref}
\usepackage{mathtools,tikz-cd}
\theoremstyle{definition}
\newtheorem*{theorem*}{Theorem}

\newtheorem*{proposition*}{Proposition}

\newtheorem*{definition*}{Definition}

\newtheorem*{lemma*}{Lemma}
\newtheorem{lemma}{Lemma}
\newtheorem*{claim*}{Claim}

\usepackage{cite}
\usepackage{amsmath,amssymb,amsfonts}
\usepackage{textcomp}
\usepackage{xcolor}

\usepackage{geometry}
\begin{document}

\title{A Prioritized Trajectory Planning Algorithm for Connected and Automated Vehicle Mandatory Lane Changes
}

\author{Nachuan Li, Austen Z. Fan, Riley Fischer, Wissam Kontar, and Bin Ran{*}}

\maketitle

\begin{abstract}

We introduce a prioritized system-optimal algorithm for mandatory lane change (MLC) behavior of connected and automated vehicles (CAV) from a dedicated lane. Our approach applies a cooperative lane change that prioritizes the decisions of lane changing vehicles which are closer to the end of the diverging zone (DZ), and optimizes the predicted total system travel time. Our experiments on synthetic data show that the proposed algorithm improves the traffic network efficiency by attaining higher speeds in the dedicated lane and earlier MLC positions while ensuring a low computational time. Our approach outperforms the traditional gap acceptance model.

\end{abstract}

\begin{IEEEkeywords}
Connected and Automated Vehicles, Dedicated Lane, Mandatory Lane Change
\end{IEEEkeywords}

\section{Introduction}
\subsection{\textbf{Overview}}
The presence of connected and automated vehicle highway (CAVH) technology has given a unique opportunity to improve mobility through vehicle-to-vehicle (V2V) and vehicle-to-infrastructure (V2I) communications\cite{ran2019connected, jin2020cloud}. In a CAVH system, there is typically a dedicated lane for CAVs that quarantines them from the HDVs \cite{ran2019connected,zhong2019dedicated, ong2018dedicating}, and a diverging zone (DZ)  for CAVs to carry out lane changes into the human driven vehicle (HDV) lane\cite{khattak2020cooperative}. This gives rise to a significant challenge of managing mandatory lane changes of CAVs in a limited length of a DZ efficiently.

Our paper proposes a prioritized system-optimal (PSO) algorithm for trajectory planning of CAVs that decide to execute a a MLC within the DZ, so that the total efficiency of the CAV dedicated lane is maximized. To specify how prioritization works, the decisions and predicted trajectory of the leading MLC CAVs act as constraints to decisions of following MLC CAVs. It will strive to increase the average speed of all vehicles in a DZ, and decrease the time and distance taken before lane change execution.  

MLC of CAVs is research-worthy because of the unpredictability of HDVs on the adjacent lane and the complicated nature of MLCs. MLCs entail the urgency posed by the routing decision of the vehicles, and generally has lower gap acceptance \cite{yang1996microscopic, ahmed1999modeling}, and is one of the main reason for congestion at highway bottlenecks.\cite{cassidy1999some,gong2016optimal} .

\subsection{\textbf{Related Works}}

A review of literature indicate that system-optimal trajectory planning of lane-changing CAVs maximizes the total efficiency of a network of vehicles, and solves the problem of decision conflicts\cite{yao2017optimizing, bevly2016lane}.

While global optimization achieves system-optimality, it is not practical because the calculation time is exponential with respect to the size of the system \cite{bennewitz2001optimizing}. On the other hand, decentralized system-optimal method is more widely studied. One approach is to mesoscopically optimize the trajectories of each cluster of vehicles or platoons \cite{8370703, manzinger2017negotiation}. For example, Manzinger et al. \cite{manzinger2017negotiation} applies reachability analysis on each vehicle and partitions the conflict zones for different vehicles. Another approach is to carry out trajectory planning in a prioritized manner. For instance, Plessen et al.\cite{plessen2016multi} treats the road as a sequence of the vehicles, prioritizing the decision making of leading vehicles, using them as constraints for followers in a uni-directional road. 

To sum up, there are abundant studies on system-optimal trajectory-planning algorithm of AVs. However, most of them fail to stress the impact of MLCs, and improve only the lane-based throughput. In addition, they typically focus on the trajectory planning algorithm of lane changes under mixed driving conditions and the more specific application of dedicated lanes is under-explored. 

The rest of the paper is organized as follows, section \ref{algo} introduces the settings, the decision making process, and a specific case study of the PSO algorithm. Section \ref{sim} simulates this algorithm and compares the results with the gap acceptance (GA) model.

\section{Algorithm \label{algo}}

This section describes the geometrical and technological settings of the study, and then provides the details of algorithm that makes trajectory decisions for all MLC CAVs in the DZ.

\subsection{\textbf{Setting} \label{settings} }
\begin{figure}
    \centering
    \includegraphics[width=0.9\linewidth]{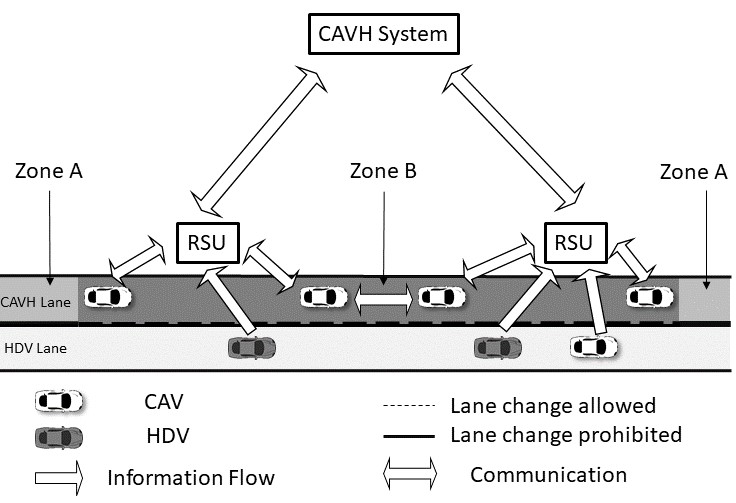}
    \caption{Geometric and Technological settings}
    \label{GTS}
\end{figure}

Fig.\ref{GTS} shows the setting of this study, which contains a CAVH dedicated lane with CAVs controlled by the system and an HDV lane with vehicles controlled manually. While the CAVH dedicated lane contains only CAVs, the HDV lane contains both CAVs and HDVs.
The arrows show the direction of information sharing. Zone A of the CAVH dedicated lane prohibits all lane changes, while zone B is the DZ.

\newgeometry{top=0.76in,bottom=0.8in,right=0.76in,left=0.76in}

Each CAV has an initial routing decision before they enter Zone A (i.e., whether to perform lane change at the current DZ) and the CAVH system carries out their MLCs based on the algorithm in section \ref{decision tree}.

In this study, we assume a CAVH system at system intelligence level 3, which means conditional automation \cite{ding2019systems}. The CAVs in the dedicated lane have V2V and V2I communication capabilities. The motion of CAVs in the CAVH dedicated lane and the lane-changing maneuvers into the HDV lane are controlled by the system. However, the CAVH system only perceives information from vehicles in the HDV lane rather than communicating with them. We also assume a traffic network with passenger cars only.

\begin{figure}
    \centering
    \includegraphics[width=0.9\linewidth]{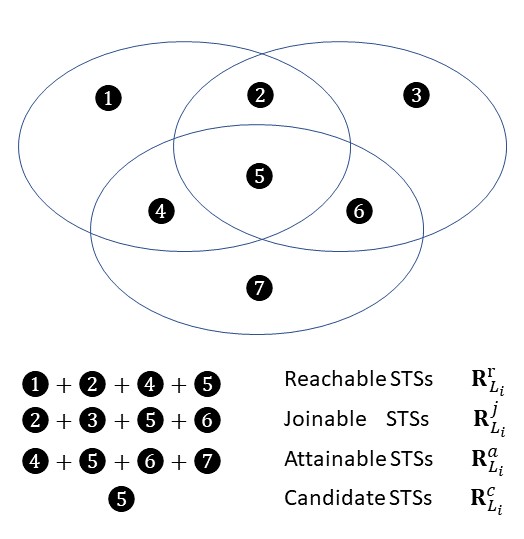}
    \caption{Relationship between Reachable, Attainable and Candidate STS's }
    \label{wien}
\end{figure}

 The process of lane change of a CAV includes three phases: (1) speed change, (2) lane change execution, (3) merge into HDV lane. We are particularly interested in phase (1) and will approximate the merging process of CAVs to be instantaneous, as also used in \cite{gong2016optimal, 8855110}. The reason for this choice is that this paper focuses on decision making of CAVs and trajectory planning before they execute MLC, which are the main challenges. Finally, we assume no system communication delay.

\subsection{\textbf{Definitions} \label{definition}}
Before discussing the PSO algorithm in details, it is necessary to define several terminologies. We start by defining a vehicle kinematic parameter (KP) as a parameter chosen by a vehicle that affects its trajectory. A sufficient tuple of KP (TKP) is the ordered set of parameters which gives the vehicle's position at each time. To be specific, the position of a vehicle is a function of time and TKP. In the rest of the paper, we treat time and KPs to be discrete variables with tiny steps.

Then, we note that a space time slot (STS) is an ordered pair of time and position\cite{manzinger2017negotiation}. We define four types of STS: (1)Reachable STS, (2)Attainable STS, (3)Joinable STS, and (4)Candidate STS. A reachable STS is an STS that could be reached by a vehicle without considering collision avoidance constraints posed by any other vehicle\cite{manzinger2017negotiation}. In other words, for each reachable STS, there exists at least one TKP that could lead to it. Define an attainable STS as an STS at which the CAV meets with minimum spacing requirement with any other vehicle in DZ, and therefore guarantees longitudinal collision avoidance. Then, define a joinable STS as an STS where a CAV is able to execute lane change because a large enough gap exists on the HDV lane and therefore guarantees lateral collision avoidance. Lastly, a candidate STS is an STS that is reachable, attainable, and joinable. Fig.\ref{wien} shows the relationship between the 4 types of STS's defined above.

Finally, we define a feasible trajectory as a trajectory whose STS's are all reachable and attainable, and ends with a candidate STS. Therefore, a feasible trajectory guarantees collision avoidance and reachability by the vehicle. 

\subsection{\textbf{Prioritized System-Optimal Decision Tree}\label{decision tree}}
The PSO decision making process provides a generalized framework for trajectory planning of MLC CAVs in a DZ under different car following models, and is divided into 5 steps: (1) Sorting and Routing Classification, (2) Reachability Analysis, (3) Attainability Analysis, (4) Joinability Analysis (5) feasible trajectory Selection. 

\begin{figure}
    \centering
    \includegraphics[width=0.75\linewidth]{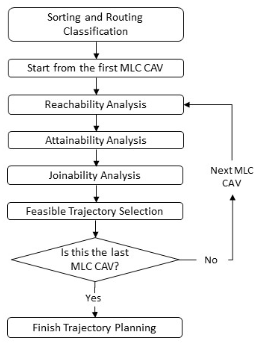}
    \caption{Prioritized System-Optimal Decision Making Process}
    \label{psd}
\end{figure}

As shown in Fig.\ref{psd}, at each time step, step (1) is carried out for all MLC CAVs in DZ simultaneously. Then the algorithm loops over all MLC CAVs in decending order for steps (2) to (5). For the decision making of each CAV, the trajectory of leading CAVs are constraints. \\ 

\subsubsection{\textbf{ Sorting and Routing Classification}}
At each time step, the decision making process starts with sorting the vehicles in the DZ based on their positions. Let $\mathbf{\Omega}:=\{\Omega_{1},\Omega_{2},...,\Omega_{N} \}$ be the sorted list of CAV's in the DZ($\Omega_{1}$ is the headmost CAV), and $\mathbf{\Omega}_L :=\{\Omega_{L_1},\Omega_{L_2},...,\Omega_{L_n} \}$ be the sorted list of all MLC CAV's among CAV's in $\mathbf{\Omega}$.  Let $\mathbf{L}:=\{L_i \mid i\in[1,n]\}$ be the set of integer indices of lane changing CAVs. 

Then, we denote the set of HDVs on the HDV lane that could affect the lane changing decisions of CAV's in set $\Omega_{L}$ as $\mathbf{\Pi}:=\{\Pi_1, \Pi_2,\dots,\Pi_{l}\}$, where $l$ is the total number of such HDV's.\\

\subsubsection{\textbf{Reachability Analysis}}
Reachability analysis obtains all reachable STS's for $\Omega_{L_i}$. 

Denote $\mathbf{Z_i}$ as the set of TKPs that could be chosen by  $\Omega_{L_i}$, and $\zeta \in \mathbf{Z_i}$ as a TKP. We can therefore define the set of reachable STS given $\zeta$ as Eq.\ref{reachable zeta} 
\begin{equation} \label{reachable zeta}
    \mathbf{R}^{r}_{L_i}\left(\zeta\right):=\{(t, x(t,\zeta)) \mid x_s\leq x(t,\zeta) \leq x_f \}
\end{equation}
where $x_{s}$ and $x_{f}$ are the starting and the ending position of DZ respectively, and $x(t, \zeta)$ is the predicted position of the vehicle at time $t$ given $\zeta$. Superposing all such $\zeta$'s, we can express the set of all reachable STS for $\Omega_{L_i}$ as Eq.\ref{Rli}\\
\begin{equation} \label{Rli}
    \mathbf{R}^{r}_{L_i}:=\underset{\zeta \in \mathbf{Z}_i}{\cup}\mathbf{R}^{r}_{L_i}\left(\zeta\right)
\end{equation}

\subsubsection{\textbf{Attainability Analysis} }
Attainability Analysis obtains all attainable STS's for $\Omega_{L_i}$. 

First, we denote the selected trajectory for vehicle $\Omega_{L_k}$ to be $J^{O}_{L_k}:=\{(t, x_{L_k}(t))\mid x_{L_k}\leq x_{L_k}(t) \leq x^{J}_{L_k}\}$ where $x^{J}_{L_k}$ is the predicted position where $\Omega_{L_k}$ executes a MLC into the HDV lane.

Then, we can denote the set of attainable STS's for $\Omega_{L_i}$ on the dedicated lane to be Eq.\ref{attainable}.
\begin{equation} \label{attainable}
    \begin{aligned}
    \mathbf{R}^{a}_{L_{i}}: =\{(t, x)\mid \forall (t', x'_{L_m}) \in \underset{m<i}{\cup} J^{O}_{{L_{m}}},  \\ t' \leq t,  x(t') \leq x'_{{L_{m}}}-M(\Omega_{L_i}, \Omega_{L_{m}}, t') \}
\end{aligned}
\end{equation}

Where $x'_{L_{m}}$ is the position of $\Omega_{L_{m}}$ at time $t'$ on the CAVH dedicated lane, $\underset{m<i}{\cup} J^{O}_{{L_{m}}}$ is the set of STS's taken by all leading MLC CAVs of $\Omega_{L_{i}}$. $M$ is a function that determines the minimum spacing between vehicle $\Omega_{L_{i}}$ and $\Omega_{L_m}$ at time $t$. This set contains STS's that ensure safe spacing of $\Omega_{L_{i}}$ with its leaders.
Details of $M$ will be discussed in section \ref{case study}\\

\subsubsection{\textbf{Joinability Analysis} }

Joinability Analysis Derives at all joinable STS's for $\Omega_{L_i}$. 

We note that there are two kinds of trajectories in the HDV lane that are in potential conflict with $\Omega_{L_{i}}$: predicted trajectory of HDVs, and leading CAVs after a MLC. Denote the set of those vehicles $\mathbf{\Pi} \cup \mathbf{\Omega}_{L_{<i}}$, where $\mathbf{\Omega}_{L_{<i}}$ denotes leading MLC CAVs in DZ of $\Omega_{L_{i}}$.

The CAVH system does not control vehicles in the HDV lane, but rather predicts their motion. We denote $\mathbf{H}_{L_{i}}:=H(\mathbf{\Pi} \cup \mathbf{\Omega}_{L_{<i}})$ as the function that derives at the predicted STS's on the HDV lane for vehicles in $\mathbf{\Pi} \cup \mathbf{\Omega}_{L_{<i}}$. Details of function $H$ will be discussed in \ref{case study}.
    
Then we denote the set of joinable STS's of $\Omega_{L_{i}}$ as Eq.\ref{joinable}.
\begin{equation} \label{joinable}
\begin{aligned}
    \mathbf{R}^j_{L_{i}}: =G(\mathbf{H}_{L_{i}})
\end{aligned}
\end{equation}
where function $G$ derives the set of joinable STS's of $\Omega_{L_{i}}$ from the predicted STS's taken by vehicles on the HDV lane. The details of the transformation $G$ will be discussed in the section \ref{case study}.\\ 

\subsubsection{\textbf{Feasible Trajectory Selection} }
Feasible Trajectory selection obtains the set of feasible trajectories of $\Omega_{L_{i}}$, and selects the optimal trajectory among them.

By taking the intersection among reachable, attainable, and joinable STS's, we can express the set of candidate STS's as Eq.\ref{candidate}
\begin{equation} \label{candidate}
\begin{aligned}
    \mathbf{R}^c_{L_{i}}: = \mathbf{R}^r_{L_{i}}\cap \mathbf{R}^a_{L_{i}}\cap \mathbf{R}^j_{L_{i}}
\end{aligned}
\end{equation}

By definition in \ref{definition}, a feasible trajectory ends with a candidate STS, and all STS's in the trajectory are both reachable under the same TKP and attainable. Therefore, we can denote $\mathbf{X}^c_{L_i}(\zeta):=\{x \mid \exists t: (t,x) \in \mathbf{R}^c_{L_{i}} \wedge (t',x'(t',\zeta)) \in \mathbf{R}^a_{L_{i}} \ \forall t'<t \}$ as the set of positions that a feasible trajectory could end with given the TKP $\zeta$. Since $\zeta$ already ensures reachability for $\Omega_{L_i}$, only attainability and joinability needs to be considered when formulating $\mathbf{X}^c_{L_i}(\zeta)$.

As defined in section.\ref{definition}, $x$ is known given a certain $t$ and $\zeta$. We can therefore express the trajectory of $\Omega_{L_i}$ given $\zeta$ and the MLC execution position $x^J$ as Eq.\ref{Jm}
\begin{equation} \label{Jm}
\begin{aligned}
    {J}_{L_{i}}(\zeta, x^J):= \underset{x<x^J}{\cup}\{(t,x(t,\zeta))\}
\end{aligned}
\end{equation}

When we superpose all TKPs $\zeta \in \mathbf{Z}_i$ and all possible MLC positions given each TKP, we can derive at the set of feasible trajectories, as shown by Eq.\ref{Jtraj} 
\begin{equation} \label{Jtraj}
    \begin{aligned}
       \mathbf{J}_{L_{i}}:=\underset{\zeta \in \mathbf{Z}_i}{\cup}\underset{x_J \in \mathbf{X}^c_{L_i}(\zeta)}{\cup}{J}_{L_{i}}(\zeta,x^J)
    \end{aligned}
    \end{equation}

\begin{figure}
    \centering
    \includegraphics[width=1\linewidth]{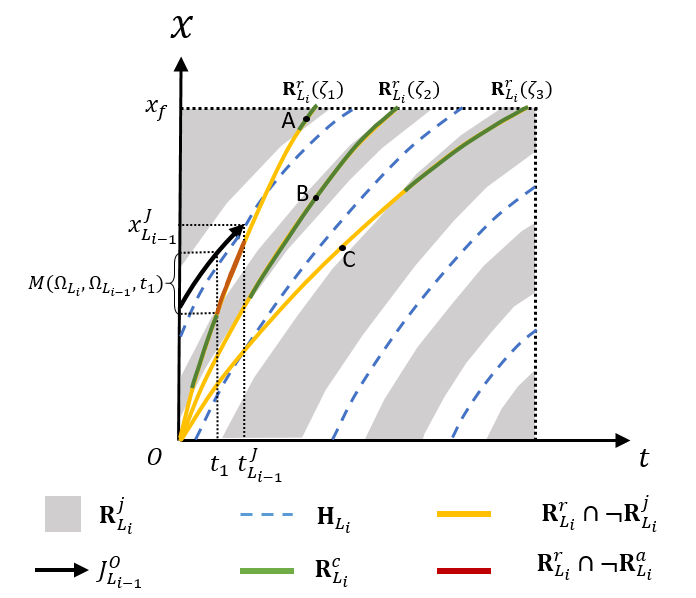}
    \caption{Feasible Trajectory Selection}
    \label{conf}
\end{figure}

Fig.\ref{conf} shows the process of feasible trajectory selection of vehicle $\Omega_{L_i}$, assuming that there are only 3 possible TKPS $\zeta_1$, $\zeta_2$, and $\zeta_3$ for $\Omega_{L_i}$. In this figure, the dashed lines indicate the predicted STS's of vehicles in the HDV lane and the shaded areas indicate the joinable STS's of $\Omega_{L_i}$. The green lines are candidate STS's, the yellow lines are STS's that are reachable but not joinable, and the red lines are STS's that are reachable but not attainable. According to the definition in section \ref{definition}, OB is a feasible trajectory, but OA is not because some of its STS's are not attainable, and OC is not because it does not end with a candidate STS.

After we obtain the set of all feasible trajectories of $\Omega_{L_{i}}$, we select the trajectory that minimizes a certain cost function.

Define $C(J_{L_{i}})$ as the cost function that depends on the trajectory $J_{L_i} \in \mathbf{J}_{L_{i}}$. Denote $J^{O}_{L_{i}}$ as the optimized trajectory that minimizes cost function $C$,  $x^{J}_{L_{i}}$ and $t^{J}_{L_{i}}$ as the position and time where $\Omega_{L_{i}}$ executes a MLC into the HDV lane respectively given $J^{O}_{L_{i}}$. Details of the cost function $C$ will be discussed with more details in section \ref{cost function}.

Algorithm \ref{alg1} summarizes general framework of the decision making process discussed in the steps above.\\

\begin{algorithm}

\caption{Path planning for all MLC CAVs in CAVH dedicated
lane DZ at each time step $t$}\label{alg1}
\begin{algorithmic}[1]
\State Sort CAVs in the dedicated lane DZ into $\mathbf{\Omega}$
\State  Sort MLC CAVs in $\mathbf{\Omega}$ into $\mathbf{\Omega}_{L}$
\State Get the set of HDVs $\mathbf{\Pi}$ that are relevant to decision making of CAVs in $\mathbf{\Omega}_{L}$
\State $i \leftarrow 1$
\While{$i \leq n$}
    \State Get the set of reachable STS's $\mathbf{R}^r_{L_{i}}$ for $\Omega_{L_i}$
    \State Get the set of attainable STS's $\mathbf{R}^{a}_{{L_{i}}}$ for $\Omega_{L_i}$
	\State Get the set of joinable STS's $\mathbf{R}^{j}_{{L_{i}}}$ for $\Omega_{L_i}$
	\State Get the set of feasible trajectories $\mathbf{J}_{L_i}$
	\State Select the trajectory $J^{O}_{L_{i}} \in \mathbf{J}_{L_i}$ for $\Omega_{L_i}$ that minimizes the cost function $C$ 
	\If{$x_{L_i}=x^{J}_{L_{i}}$}
	    \State $\Omega_{L_i}$ execute lane change
	\EndIf 
	\State $i \leftarrow i+1$
\EndWhile

\end{algorithmic}
\end{algorithm}

\subsection{\textbf{Case Study} \label{case study}}
In order to explain the mechanism of our PSO algorithm discussed in section \ref{decision tree}, we propose a case study with a cost function and explore its application in a spring mass damper car following model for CAVs and uniform speed prediction on vehicles in the HDV lane.\\

\subsubsection{\textbf{Cost Function} \label{cost function}}

We propose a time-based cost function that guides the decision making process in section \ref{algo}. 

Before the MLC into the HDV lane by CAV $\Omega_{L_i}$, its speed change can influence its followers, and can therefore lead to a change in total travel time of CAVs in the DZ. Also, we assume that the purpose of a CAV to perform a MLC is to exit the highway, and needs to perform a second lane change into the shoulder lane directly attached to the off ramp. Therefore, there exists a risk that $\Omega_{L_{i}}$ fails to get to the off ramp, and thus leading to a detour. With this information, we model the cost function as the sum of (1) Expected detour time of $\Omega_{L_i}$, and (2) Expected total delay in DZ, shown by Eq.\ref{tot_cost}.

\begin{equation} \label{tot_cost}
    C^{time}_{L_{i}}:=e^{-0.046\left(x_f-x^{J}_{L_{i}}\right)} \frac{X_d}{v_d} + N^{f}_{L_{i}} \left(t^J_{L_i}- \frac{x^{J}_{L_{i}}-x_{L_{i}}}{v_{L_i}} \right)
\end{equation}
and the constraints as Eq.\ref{const1} and Eq.\ref{const2}
\begin{equation} \label{const1}
  \left(t^J_{L_i},x^{J}_{L_{i}} \right) \in \mathbf{R}^j_{L_i}
\end{equation}
\begin{equation} \label{const2}
    \exists  \textbf{ } J_{L_i}\in \mathbf{J}_{L_i} \text{ such that } \left(t^J_{L_i},x^{J}_{L_{i}} \right) \in J_{L_i}
\end{equation}

The first term $e^{-0.046\left(x_f-x^{J}_{L_{i}}\right)} \frac{X_d}{v_d} $ represents the expected detour time of $\Omega_{L_i}$, where $X_d$ is the detour distance if $\Omega_{L_{i}}$ fails to leave the highway and $v_d$ is the estimated detour speed. $e^{-0.046\left(x_f-x^{J}_{L_{i}}\right)}$ is the failure rate to carry out a second MLC into the offramp\cite{cao2018analysing} . The failure rate increases as the position to execute a MLC $x^{J}_{L_{i}}$ increases and approaches 1 if the vehicle executes a MLC at the end of DZ. 

The second term $N^{f}_{L_{i}} \left(t^J_{L_i}- \frac{x^{J}_{L_{i}}-x_{L_{i}}}{v_{L_i}} \right)$ is the expected total delay in the DZ due to $\Omega_{L_i}$'s speed change. $t^{J}_{L_{i}}$ and $x^{J}_{L_{i}}$ are the predicted time and position where $\Omega_{L_i}$ executes MLC respectively. $N^{f}_{L_{i}}$ is the number of followers of $\Omega_{L_i}$, and $v_{L_i}$ is the current speed of $\Omega_{L_i}$. 

Finally, we show that the constraints in Eq.\ref{const1} and Eq.\ref{const2} are sufficient:

Eq.\ref{const1} ensures that the STS's, where $\Omega_{L_i}$ executes a MLC, are able to join. Eq.\ref{const2} ensures that the STS's are on at least one feasible trajectory. By definition in section \ref{definition}, all STS's on a feasible trajectory are reachable and attainable, which meet with kinematic and spacing constraints. Therefore, the constraints are sufficient.\\

\subsubsection{\textbf{Car Following Model}}
For the case study, we use a spring mass damper car following model. This model has been widely studied and applied in similar settings in literature about  AVs\cite{bang2017platooning,jiang2020risk, rao2019analysis}. In this model, the acceleration of a vehicle is expressed as Eq.\ref{mass_damper}  .
\begin{equation} \label{mass_damper}
   \ddot{x}=\alpha(\Delta x-\tilde{s}) - \beta( \dot{x} -\tilde{v})
\end{equation}
where $\ddot{x}$, $\dot{x}$ are the acceleration and speed respectively. $\Delta x$ is the spacing between the vehicle and its leader, $\tilde{s}$ is the desired inter-vehicle spacing, and $\tilde{v}$ is the desired speed. Furthermore, $\alpha$ and $\beta$ are coefficients determining the sensitivity of acceleration with respect to the vehicle's deviation from the desired spacing and speed respectively. For the CAVs, a larger $\alpha$ means higher connectedness between vehicles, and a larger $\beta$ means stronger speed control.

For a lane changing CAV $\Omega_{L_{i}}$, we set $\alpha=0$, since it will lose connection with its leader when it executes MLC. Its acceleration can therefore be modeled as Eq.\ref{mass_damper_Li} 
\begin{equation} \label{mass_damper_Li}
   \ddot{x}_{{L_i}}(t,\beta_{{L_{i}}})= -\beta_{L_{i}}( \dot{x}_{{L_i}}(t) -v_{min})
\end{equation}
where $v_{min}$ denotes the minimum speed of CAVH dedicated lane and $\beta_{L_{i}}$ denotes the KP of $\Omega_{L_{i}}$ such that $0\leq\beta_{L_{i}}\leq\beta_{max}$. $\beta_{L_{i}}=0$ and $\beta_{L_{i}}=\beta_{max}$ denote the constant speed and the largest deceleration scenario respectively. 

Integrating of Eq.\ref{mass_damper_Li} produces the relationship between the lane changing execution time $t^J_{L_i}$, the lane changing position $x^{J}_{L_{i}}$, and $\beta_{L_{i}}$, as Eq.\ref{merge_x}.

\begin{equation} \label{merge_x}
    x^J_{L_i}= -\frac{v_{{L_i}}-v_{min}}{\beta_{L_{i}}}e^{\beta_{L_{i}} t^J_{L_i}}+v_{min}t^J_{L_i}
    \\
    +x_{L_{i}}+\frac{v_{{Li}}-v_{min}}{\beta_{{L_{i}}}}
\end{equation}
where $v_{{Li}}$ is the initial velocity of $\Omega_{L_{i}}$.\\

In this case study, $\beta_{L_{i}}$ is the only KP of $\Omega_{L_{i}}$. Therefore the set of possible TKPs can be written as $\mathbf{Z}_i=\{\{\beta_{L_{i}}\} \mid 0 \leq \beta_{L_{i}} \leq \beta_{max} \}$\\

\subsubsection{\textbf{Speed Prediction Model}}

We predict the speed of vehicles in the HDV lane with uniform speed prediction, as shown by Eq.\ref{spm}
\begin{equation} \label{spm}
    x=x_{\Pi}+v_{\Pi}t
\end{equation}
where $\Pi$ denotes a vehicle in the HDV lane, $x_{\Pi}$ and $v_{\Pi}$ are its current position and speed respectively. We use  uniform speed prediction model for its simplicity.\\

\subsubsection{\textbf{Function Definitions}}
Here we define the functions $M$, $H$, and $G$ as mentioned in section \ref{decision tree} specifically for our case study.

Recall that $M\left(\Omega_{L_{i}},\Omega_{L_{m}},t \right)$ is a function that determines the minimum spacing between two MLC CAVs $\Omega_{L_{i}}$ and $\Omega_{L_{m}}$ at time $t$. In this case study, we define $M$ as Eq.\ref{M}
\begin{equation} \label{M}
M\left(\Omega_{L_{i}},\Omega_{L_{m}},t \right) := \left(L_{i}-L_{m}\right) h^{CAV}_{min}v^{CAV}_{max}
\end{equation}
where $L_{i}-L_{m}$ is the number of spacings between $\Omega_{L_{i}}$ and $\Omega_{L_{m}}$ . Also, $h^{CAV}_{min}$ and $v^{CAV}_{max}$ are the minimum headway and the maximum speed on CAVH dedicated lane respectively. The expression $h^{CAV}_{min}v^{CAV}_{max}$ is the minimum safe spacing given the highest possible speed on the CAVH dedicated lane, and would therefore guarantee one safe inter-vehicle spacing when the exact speed of CAVs are unknown. 

Recall that $H\left(\mathbf{\Pi}\cup \mathbf{\Omega}_{L_{<i}}\right)$ is the function that determines the predicted STS's for vehicles in $\mathbf{\Pi}\cup \mathbf{\Omega}_{L_{<i}}$ in the future. Since we predict vehicles in HDV lane to move with uniform speed, we define $H$ as Eq.\ref{H}
\begin{equation} \label{H}
\begin{split}
 H(\mathbf{\Pi}\cup \mathbf{\Omega}_{L_{<i}}) :=
\underset{{\Pi} \in \mathbf{\Pi}\cup \mathbf{\Omega}_{L_{<i}}}{\cup} 
\{(t,x) \mid x=x_{\Pi}+v_{\Pi}(t-t^{J}_{\Pi}) \\ 
\text{ such that } 
x_s-h^{HDV}_{{min}}v^{HDV}_{{max}} \leq x \leq x_f+h^{HDV}_{{min}}v^{HDV}_{{max}} \}
\end{split}
\end{equation}

If $\Pi$ is an HDV, then $x_{\Pi}$ and $v_{\Pi}$ are the current longitudinal position and velocity of $\Pi$ on the HDV lane respectively, and $t^J_{\Pi}$ is 0. If $\Pi$ is a CAV, then $x_{\Pi}$ and $v_{\Pi}$ are the predicted position and velocity of $\Pi$ when it executes a MLC into the HDV lane, and $t^J_{\Pi}$ is its predicted time to execute a MLC.

Note that for all leading lane-changing CAVs of $\Omega_{L_i}$, the spacing on the HDV lane is reserved by placing a virtual vehicle on the HDV before its lane change, expressed by $x_{\Pi}+v_{\Pi}(t-t^{J}_{\Pi})$ for $t<t^{J}_{\Pi}$.  $h^{HDV}_{{min}}$ denotes the minimum headway on the HDV lane, and $v^{HDV}_{{max}}$ denotes the maximum headway on the HDV lane. Therefore, $h^{HDV}_{{min}}v^{HDV}_{{max}}$ guarantees GA for $\Pi$. Recall that $x_s$ and $x_f$ are the starting and ending position of DZ, we note that $x_s-h^{HDV}_{{min}}v^{HDV}_{{max}}$ and $x_f+h^{HDV}_{{min}}v^{HDV}_{{max}}$ are lower and upper bounds of longitudinal positions for an HDV to possibly influence the decision making of an MLC CAV.

Recall that $G\left(\mathbf{H}_{L_{i}}\right)$ is the transformation from all predicted STS's taken on the HDV lane relative to the lane-changing decision of vehicle $\Omega_{L_i}$ to joinable STS's. In this case study,  $G$ is defined as Eq.\ref{G}
\begin{equation} \label{G}
\begin{split}
    & G\left(\mathbf{H}_{L_{i}}\right):= \\
    & \{(t,x) \mid \forall (t,x') \in \mathbf{H}_{L_{i}},  |x-x'| \geq h^{HDV}_{{min}}v^{HDV}_{{max}}  \}
\end{split}
\end{equation}
The inequality  $|x-x'| \geq h^{HDV}_{{min}}v^{HDV}_{{max}}$ guarantees a safe spacing for all joinable STS's.\\

\subsubsection{\textbf{Cost Minimization Algorithm} \label{cfa}}
Here, we derive the cost minimization algorithm that selects the optimal trajectory of each MLC CAV at each time step. 

From Eq.\ref{merge_x} and Eq.\ref{tot_cost}, we derive at lemma \ref{l1}-lemma \ref{bt}. These lemmas help produce a simpler algorithm for the case study by eliminating some obvious non-optimal solutions.

\begin{lemma}\label{l1}
For each MLC CAV $\Omega_{L_{i}}$, given a fixed travel time $t^{J}_{{L_{i}}}$, a larger KP $\beta_{{L_i}}$ results in a lower distance travelled $x^{J}_{{L_{i}}}$.
\end{lemma}
\begin{proof}
Based on Eq.\ref{merge_x}, taking the partial derivative of $x^{J}_{{L_{i}}}$ with respect to $\beta_{{L_{i}}}$ is always non-positive given constant $t^J_{_{L_i}}$.
\end{proof}
\begin{lemma}\label{l2}
For each MLC CAV $\Omega_{L_{i}}$, given a constant KP $\beta_{{L_i}}$, a lower time $t^J_{{L_{i}}}$ taken before MLC execution results a lower total cost $C^{time}_{L_i}$.
\end{lemma}

\begin{proof}
While treating $t^J_{{L_{i}}}$ and $\beta_{{L_i}}$ as independent variables, the partial derivative of total time cost in Eq.\ref{tot_cost} with respect to $t^J_{{L_{i}}}$ is always non-negative. 
\end{proof}

\begin{lemma}\label{l3}
For each MLC CAV $\Omega_{L_{i}}$, given a MLC position $x^{J}_{{L_i}}$, a lower time $t^J_{{L_{i}}}$ taken before MLC execution produces a lower total cost $C^{time}_{L_i}$.
\end{lemma}
\begin{proof}
While treating $t^J_{L_{i}}$ and $x^{J}_{L_i}$ as independent variables, the partial derivative of total time cost in Eq.\ref{tot_cost} with respect to $t^J_{L_{i}}$ is always non-negative.
\end{proof}

\begin{lemma}\label{immed}
If the current position of $\Omega_{L_{i}}$ is in the DZ and joinable, performing MLC immediately optimizes the trajectory.

\end{lemma}

\begin{proof}
This is a direct corollary of lemma \ref{l2}.
\end{proof}

\begin{lemma}\label{bt}
Given two STS $S_1=(t_1, x(t_1,\beta_1))$ and $S_2=(t_2, x(t_2,\beta_2))$ on feasible trajectories such that $t_1>t_2$, $\beta_1>\beta_2$, and $x(t_1,\beta_1)>x(t_2,\beta_2)$, executing a MLC at $S_2$ produces a lower cost than at $S_1$.

\end{lemma}

\begin{proof}
Take an STS $S_3=(t_3, x(t_3, \beta_2))$ such that\\ $x(t_3, \beta_2)=x(t_1, \beta_1)$. Since $\beta_1>\beta_2$, we have $x(t_1, \beta_2)<x(t_3, \beta_2)$, based on lemma \ref{l1} and $x(t_1, \beta_2)<x(t_1, \beta_1)$. Since the speed is always positive, $t_3>t_1$ is satisfied. Since the positions of $S_3$ and $S_1$ are equal, and $t_3>t_1$, executing a MLC at $S_1$ yields a higher cost than at $S_3$ based on lemma \ref{l3}. Now since $t_1>t_2$, we have $t_3>t_2$. Compare $S_2$ and $S_3$, they have the same KP $\beta_2$ and $t_3>t_2$. This implies, according to lemma \ref{l2}, executing a MLC at $S_3$ yields a higher cost than at $S_2$. In summary, executing a MLC at $S_2$ yields a lower cost than at $S_1$. 
\end{proof}

Based on the lemmas above, algorithm \ref{alg2} selects the optimal trajectory for $\Omega_{L_i}$. $\Delta t$ and $\Delta \beta$ are the minimum increments of time $t$ and KP $\beta$. Based on lemma \ref{immed}, if the current position is joinable, the MLC CAV performs lane change immediately. Otherwise, values of $\beta$ and MLC execution time step $t$ are looped over. Based on lemma \ref{l2}, for each $\beta$, we only consider the smallest $t$ that produces a feasible trajectory. Based on lemma \ref{bt}, if a KP $\beta$ is larger than another KP that could produces a feasible trajectory at some time $t^J_{L_i}$, then only $t$ that is lower than all locally optimal merging time for smaller KPs are considered.

By applying algorithm \ref{alg2} into the feasible trajectory selection part from algorithm \ref{alg1}, we get the trajectory of each MLC CAV that minimizes the total cost shown in Eq.\ref{tot_cost}, under constraints in Eq.\ref{const1} and Eq.\ref{const2}

\begin{algorithm}
\caption{Trajectory optimization for a MLC CAV $\Omega_{L_{i}}$  }\label{alg2}
\begin{algorithmic}[1]

\If{$x_{L_i} \in \mathbf{R}^j_{L_{i}}$}
    \State $x^{J}_{L_{i}} \leftarrow x_{L_{i}} $
    , $\beta_{L_{i}} \leftarrow 0 $
    , $t^{J}_{L_{i}} \leftarrow 0 $

\Else
    \State $x^{J}_{L_{i}} \leftarrow$ None
    , $t^{J}_{L_{i}} \leftarrow \frac{x_f-x^{J}_{L_{i}}}{v_{min}} $
    , $C^{time}_{L_{i}} \leftarrow \infty$
    , $\beta_{L_{i}} \leftarrow 0 $
\EndIf
\While{$\beta < \beta_{max}$}
    \State $\beta \leftarrow \beta + \Delta \beta$
    \While{$t < t^{J}_{L_{i}}$}
        \State $t \leftarrow t+\Delta t$
        \If{$x(t,\beta) \in \mathbf{R}^{c}_{L_{i}}$ and $C^{time}_{L_{i}}(J_{L_i}(\{\beta\}, x(t,\beta))) < C^{time}_{L_{i}}$}
            \State $x^{J}_{L_{i}} \leftarrow x(t, \beta)$
            , $t^{J}_{L_{i}} \leftarrow t$
            , $\beta_{L_{i}} \leftarrow \beta$
            \State $C^{time}_{L_{i}} \leftarrow C^{time}_{L_{i}}(J_{L_i}(\{\beta\}, x(t,\beta)))$
            \State \textbf{break}
        \EndIf
    \EndWhile
\EndWhile

\end{algorithmic}
\end{algorithm}

\section{{Simulation} \label{sim}} 

This section simulates the PSO decision tree of MLC CAVs in DZ for the case study in section \ref{case study} with multiple vehicles. The simulation will compare the decision making process in this paper and the widely applied GA model.

\subsection{\textbf{Settings}}
This simulation is carried out on a straight segment of highway with an inner CAVH dedicated lane, and an outer HDV lane, as shown in Fig.\ref{GTS}. DZ starts at $x_s=0$m and ends at $x_f=1500$m, between which MLC CAVs change speed and perform MLCs into the HDV lane. The simulation is carried out discretely with a time step of 0.2s. In each time step, MLC CAVs make their trajectory decisions and each vehicle updates its speed and position.

\subsection{\textbf{Models}}
Vehicles in HDV lane use Newell Car following model\cite{newell2002simplified, chen2012behavioral}, with traffic wave speed 3.7m/s and average jam spacing 3.7m \cite{newell2002simplified}. CAVs in DZ $\Omega_{L_{i}}$  use the  spring mass damper model shown by Eq.\ref{merge_x}.

We assume that the minimum headway on the HDV and CAVH dedicated lane are $h^{HDV}_{min}=1.5$s and  $h^{CAV}_{min}$=0.5s \cite{948767, seraj2018modeling}. The maximum deceleration and acceleration are $-4$m/s$^2$ and $2$m/s$^2$ respectively \cite{bokare2017acceleration}. We therefore express the maximum value of $\beta_{L_{i}}$ as $\frac{72}{5(v_{{L_{i}}}-v_{min})}$, where $v_{{L_{i}}}$ is the current speed of vehicle $\Omega_{L_{i}}$ and $v_{min}$ is the minimum speed on the CAVH dedicated lane, both in units of km/h.\\

\subsubsection{\textbf{Vehicle Generation}}
We initialize with 5 MLC CAVs in a DZ with speed 100km/hr, and 8 HDVs on HDV lane with speed between 60km/hr and 100km/hr. The desired speed of all vehicles in HDV lane are between 80km/hr and 100km/hr. A vehicle's desired speed is its optimal speed on the HDV lane under no inter-vehicle interactions. \\ 

\subsubsection{\textbf{Gap Acceptance Model}}
Gap acceptance (GA) model is widely used for MLCs near freeway bottlenecks. The following lead to a smaller GA: A position closer to the latest possible MLC (i.e. \ $x_f$), a lower speed, and a higher density on the target lane \cite{ali2018connectivity, hao2020research, sun2012lane}. Eq.\ref{ga} models the minimum headway of MLCs, as proposed in \cite{hao2020research}.
\begin{equation} \label{ga}
    h_{MLC}=\frac{x_f-x}{x_f-x_s} h^{HDV}_{min}
\end{equation}
$h_{MLC}$ denotes the minimum headway during a MLC, $x$ denotes the current position of the vehicle, and $h^{HDV}_{min}$ denotes the minimum headway in the HDV lane under general environment. Eq.\ref{ga} shows that the MLC headway decreases as the CAV approaches the end of DZ.

Under GA model, the MLC CAVs decelerates gradually to the desired speed of the HDV lane.\cite{fransson2018driving}.\\

\begin{figure*}[t!]
  \centering
  \begin{subfigure}[b]{0.4\linewidth}
  \begin{center}
    \includegraphics[width=1\linewidth]{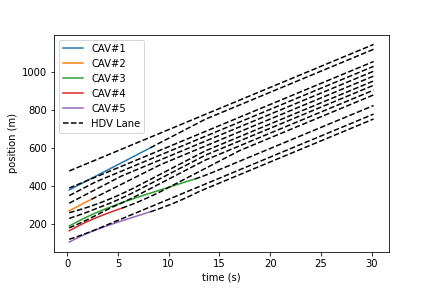}
    \caption{Trajectory of MLC CAVs and HDVs with PSO algorithm}
    \end{center}
  \end{subfigure}
  \begin{subfigure}[b]{0.4\linewidth}
    \begin{center}
    \includegraphics[width=1\linewidth]{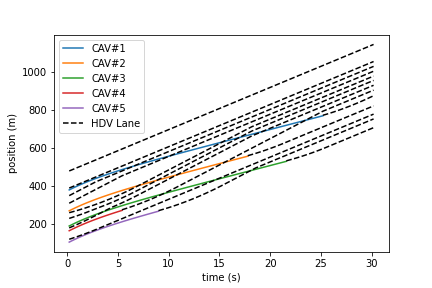}
    \caption{Trajectory of MLC CAVs and HDVs with gap acceptance model }
      \end{center}
  \end{subfigure}
  
  \begin{subfigure}[b]{0.4\linewidth}
  \begin{center}
    \includegraphics[width=1\linewidth]{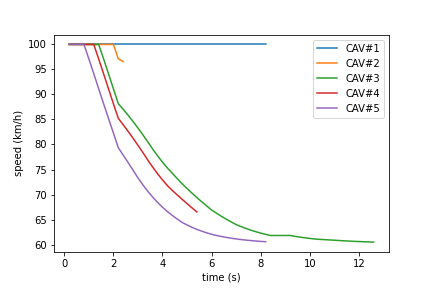}
    \caption{Speed of MLC CAVs with PSO algorithm}
      \end{center}
  \end{subfigure}
  \begin{subfigure}[b]{0.4\linewidth}
  \begin{center}
    \includegraphics[width=1\linewidth]{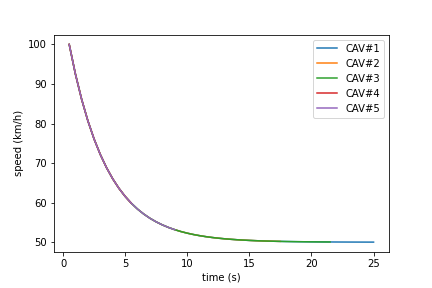}
    \caption{Speed of MLC CAVs with gap acceptance model}
    \end{center}
  \end{subfigure}
    
    \caption{Simulation Result Comparison of PSO Algorithm and GA Model }
    \label{macro}
\end{figure*}

\subsubsection{\textbf{Results}}
Fig.\ref{macro}(a) and Fig.\ref{macro}(b) shows the trajectory of all vehicles in the simulation network with PSO algorithm and GA model respectively. The dashed lines and solid lines denote the trajectory of vehicles in the HDV lane and CAVH dedicated lane respectively. The connecting point of a solid line and a dashed line denote the STS where a CAV executes MLC. Fig.\ref{macro}(c) and Fig.\ref{macro}(d) shows the speed of the lane changing CAVs in the CAVH dedicated lane before their MLC execution into the HDV lane under PSO algorithm and GA model. 

Comparing Fig.\ref{macro}(a) with Fig.\ref{macro}(b), and Fig.\ref{macro}(c) with Fig.\ref{macro}(d), we find that under PSO algorithm, the CAVs undergo relatively slower and non-simultaneous deceleration than GA model. Based on Fig.\ref{macro}(c), while the rate of deceleration of CAVs fluctuate, PSO algorithm still produces relatively smooth speed change. CAV\#1 and CAV\#2 fail to merge into a gap just slightly in front under GA model. However, they successfully merge into the closest gaps by keeping at their current speed with PSO algorithm. 

CAV\#5 gives up joining the gap before HDV\#8 under PSO algorithm, and decides to decelerate and merge into a later gap because of constraints of CAV\#3 and CAV\#4's trajectory decisions.

We also find that the average speed of the HDV lane with the PSO algorithm and GA model are both around 82km/hr. But the average speed of the DZ has increased from 63km/hr to 85km/hr when PSO algorithm is applied. The average MLC execution position and time for the GA model is $257.82$m and $15.6$s respectively, and $162.1$m and $7.16$s for the PSO algorithm respectively.\\

\section{Conclusion}
\subsection{\textbf{Research Summary}}
In this paper, a prioritized system-optimal cooperative trajectory planning algorithm (PSO) for MLCs of CAVH dedicated lane is proposed. The speed change and MLC execution position decisions are made from the MLC CAV that is closest to the end of DZ, with the decisions of leading CAVs and vehicles in HDV lane as constraints. By deriving at the reachable, joinable, and attainable STS's and analysing a customized system cost function, we select each MLC CAV's optimal trajectory.

Simulation results indicates relatively low vulnerability of unpredictable speed on the HDV lane on the decision making process, and relatively smooth trajectories of the lane changing CAVs. Compared to traditional gap acceptance model, our PSO algorithm shows a more efficient utilization of spacing on the HDV lane, produces a higher average speed in the CAVH dedicated lane and relatively early time and position for MLC CAVs to execute lane change.

\subsection{\textbf{Applications and Future Research}}

One major benefit of our PSO algorithm is its real-time performance. The run-time for MLC decision evaluation for each vehicle is 0.03s, which is relatively low for the complexity of the system. The run-time could be further improved by applying parallel computing during implementation.

A limitation of our research is that we have only considered the spring mass damper car following model, the uniform speed prediction model, and the time-based cost functions. In the next step, we will examine the implementations of our PSO algorithm on other car following models and cost functions to verify the stability of decision making. We will also implement trajectory prediction on HDVs with machine learning to further improve the decision making efficiency.

\section{Acknowledgements}
The authors sincerely thank Ron Yang, Chaolun Xu, Yushun Chen, and Yang Cheng for their kind support and helpful discussions.

\bibliographystyle{abbrv}
\bibliography{IEEE_ITSC}

\end{document}